\documentclass[journal,transmag]{IEEEtran}
\ifCLASSINFOpdf
\else
\fi
\usepackage{booktabs}       

\usepackage{hyperref}       
\usepackage{url}            
\usepackage{nicefrac}       
\usepackage{microtype}      
\usepackage{graphicx} 
\usepackage{amsfonts} 
\usepackage{amsmath}
\usepackage{amssymb}
\usepackage[capitalize]{cleveref} 
\newcommand{\unit}[1]{\ensuremath{\, \mathrm{#1}}} 

\usepackage{dsfont} 
\usepackage[font=small,skip=0pt]{subcaption} 
\usepackage{balance}
\usepackage[font=small]{caption}
\usepackage{bm} 
\usepackage{wrapfig}
\usepackage{float} 
\usepackage{multirow}

\usepackage{color}
\usepackage[svgnames]{xcolor} \definecolor{DarkGreen}{rgb}{0,0.5,0}
\definecolor{DarkRed}{rgb}{0.75,0,0}

\usepackage{tikz,mathtools}
\crefformat{equation}{(#2#1#3)}
\Crefformat{equation}{Equation~(#2#1#3)}
\Crefname{equation}{Equation}{Equations}

\usetikzlibrary{shapes,positioning,automata,arrows,fit,backgrounds,calc}
\tikzstyle{block} = [draw, fill=blue!20, rectangle,minimum height=1em,
minimum width=2em] \tikzstyle{sum} = [draw, fill=blue!20, circle, node
distance=1cm] \tikzstyle{input} = [coordinate] \tikzstyle{output} =
[coordinate] \tikzstyle{pinstyle} = [pin edge={to-,thin,black}]
\usetikzlibrary{trees} \usetikzlibrary{decorations.pathmorphing}
\usetikzlibrary{decorations.markings}
\definecolor{darkgreen}{rgb}{0,0.5,0}
\definecolor{darkred}{rgb}{220,20,60}

\DeclareMathOperator{\R}{\mathbb{R}} 

\newcommand{\first}[1]{\textbf{#1}}

\newcommand{\brackets}[1]{\left( #1 \right)}

\newcommand{\cmmnt}[1]{\ignorespaces}

\newcommand{\bit}{\begin{itemize}}
\newcommand{\ei}{\end{itemize}}

\newcommand{\graphino}{$\textit{Graphi{\~n}o}$}

\makeatletter
\renewcommand\paragraph{\@startsection{subsubsection}{4}{\z@}%
{0.25ex \@plus.5ex \@minus.2ex}%
{-.15em}%
{\normalfont\normalsize\itshape}}
\makeatother

\usepackage[authormarkuptext=name,addedmarkup=bf,authormarkupposition=left]{changes}
\definechangesauthor[name={X.~X.}, color={orange}]{xx}
\definechangesauthor[name={Y.~Y.}, color={blue}]{yy}
\definechangesauthor[name={SRC}, color={brown}]{salva}
\definechangesauthor[name={B.~L.}, color={MediumSeaGreen}]{bl}
\begin{document}
%
\title{The World as a Graph:\\ Improving El Niño Forecasts with Graph Neural Networks}


\author{\IEEEauthorblockN{Salva R{\"u}hling Cachay\IEEEauthorrefmark{1},
Emma Erickson$^*$\IEEEauthorrefmark{2},
\\
Arthur Fender C. Bucker$^*$\IEEEauthorrefmark{3, 4}, 
Ernest Pokropek$^*$\IEEEauthorrefmark{5}, 
Willa Potosnak$^*$\footnote{Contributed equally as second authors.}\IEEEauthorrefmark{6},
\\
Suyash Bire\IEEEauthorrefmark{8},
Salomey Osei\IEEEauthorrefmark{7}, and 
Bj{\"o}rn L{\"u}tjens\IEEEauthorrefmark{8}
}
\\
\IEEEauthorblockA{\IEEEauthorrefmark{1}Technical University of Darmstadt, \IEEEauthorrefmark{2}University of Illinois at Urbana-Champaign,}
\IEEEauthorblockA{\IEEEauthorrefmark{3}University of São Paulo, \IEEEauthorrefmark{4}
Technical University of Munich, \IEEEauthorrefmark{5}Warsaw University of Technology, }
\IEEEauthorblockA{\IEEEauthorrefmark{6}Duquesne University, \IEEEauthorrefmark{7}African Institute for Mathematical Sciences, \IEEEauthorrefmark{8}Massachusetts Institute of Technology}
\thanks{Manuscript received March 12, 2021. This work has been submitted to the IEEE for possible publication. Copyright may be transferred without notice, after which this version may no longer be accessible. 
\emph{(Emma Erickson, Arthur Fender C. Bucker, Ernest Pokropek, and Willa Potosnak contributed equally.) Corresponding author: S. R. Cachay (email: salvaruehling@gmail.com).}}}

\markboth{(IN REVIEW)}%
{Shell \MakeLowercase{\textit{et al.}}: Bare Demo of IEEEtran.cls for IEEE Transactions on Neural Networks and Learning Systems Journals}
%



\maketitle

\begin{abstract}
Deep learning-based models have recently outperformed state-of-the-art seasonal forecasting models, such as for predicting 
El Ni\~no-Southern Oscillation (ENSO). 
However, current deep learning models are based on convolutional neural networks which are difficult to interpret and can fail to model large-scale atmospheric patterns. In comparison, graph neural networks (GNNs) are capable of modeling large-scale spatial dependencies and are more interpretable due to the explicit modeling of information flow through edge connections.
We propose the first application of graph neural networks to seasonal forecasting.
We design a novel graph connectivity learning module that enables our GNN model to learn large-scale spatial interactions jointly with the actual ENSO forecasting task.
Our model, \graphino, outperforms state-of-the-art deep learning-based
models for forecasts up to six months ahead.
Additionally, we show that our model is more interpretable as it learns sensible connectivity structures that correlate with the ENSO anomaly pattern.
\end{abstract}



\IEEEdisplaynontitleabstractindextext

%
\IEEEpeerreviewmaketitle


\renewcommand{\thesection}{\Roman{section}}
\section{Introduction}\label{sec:intro}
\begin{figure*}
 \centering
 \begin{subfigure}{0.9\textwidth}
  \centering
        \includegraphics[width=\textwidth]{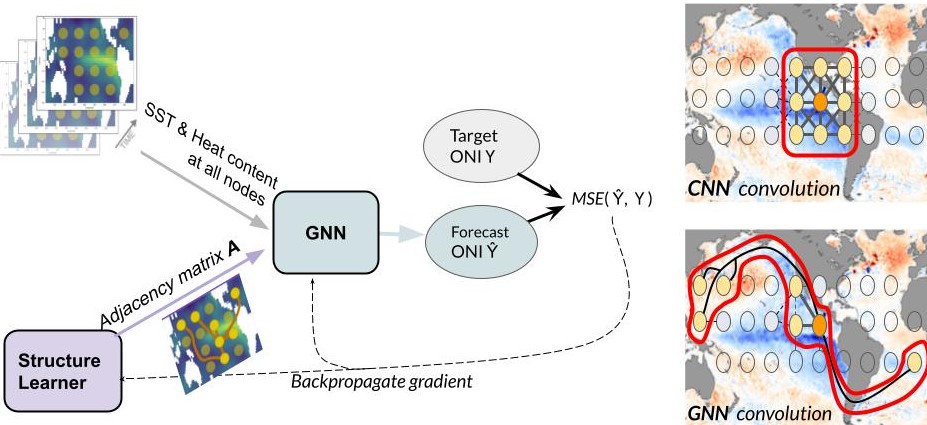}
  \end{subfigure}
\caption[teaserfigure]{We propose Graph Neural Networks (GNNs) to forecast El Niño–Southern Oscillation (ENSO).
GNNs can extract patterns at a global scale indicative of ENSO, contrary to CNNs, which are based on spatially local feature extractors, i.e. grid convolutions, and assume translational equivariance. The right part of the figure visualizes this key difference in a toy example. In this work, the goal is forecast the ONI, which is the averaged sea surface temperature anomalies over the ONI region (5$^\circ$N-5$^\circ$S, 120$^\circ$-170$^\circ$W) over three months. We jointly learn a global graph connectivity structure, represented by an adjacency matrix, with our proposed structure learning module.
} 
\label{fig:teaser} 
\end{figure*}
\IEEEPARstart{E}{l} Niño–Southern Oscillation (ENSO)
has a large influence on climate variability as it causes disasters such as floods~\cite{ENSO_floods}, droughts~\cite{ENSO_droughts}, and heavy rains~\cite{ENSO_hurricanes, ENSO_precipitation} in various regions of the world.
It also has severe implications on public health \cite{ENSO_health, ENSO_health2}.
ENSO forecasts have remained at traditionally low skill due to the high variability of ENSO manifestations and the difficulty in capturing the global scale and complexity of the ocean-atmosphere interactions that cause it~\cite{ENSO_Review2}.
Data-driven forecasting systems are additionally confronted by the limited availability of observational data.
Various indices exist to measure the presence and strength of ENSO events. As in the most related work~\cite{CNN_ENSO}, we here focus on forecasting the commonly used Oceanic Ni\~no Index (ONI).
In a recent work, a deep learning system based on a convolutional neural network 
(CNN)  was successfully applied~\cite{CNN_ENSO} to  forecasting ENSO by exploiting vast amounts of simulation data from climate models~\cite{cmip5}.
The CNN model was indeed able to outperform state-of-the-art dynamical systems, and provide skillful forecasts of the ONI for up to 17 months ahead.
However, some of the fundamental assumptions behind CNNs~\cite{DL_book} are not well suited for seasonal and long range forecasting:
\begin{itemize}
    \item 
     Parameter-shared convolutions lead to \emph{translational equivariance}, meaning that if the input is moved, its output representation will move by the same amount. In earth science applications however, the location of a certain pattern is very important. For example, sea surface temperature anomalies occurring in the tropical Pacific should be treated differently from those occurring in the north Atlantic. 
    \item
     CNNs build representations from spatially close regions of the input, leading to a \emph{spatial locality bias}. Many climate phenomena however, are driven by global interactions. 
     CNNs account for large-scale patterns only through deep layers, which misses the importance of modeling predominantly large-scale patterns.
    \item
     CNNs need to use all grid cells of the input. This makes them inflexible in cases where some regions of the input are known to not be needed and must be masked out for the CNN (e.g., as in this work, all terrestrial locations could be discarded when only using oceanic variables).  
\end{itemize}

Therefore, we advocate for formulating the \emph{ONI forecasting problem as a graph regression problem, and model it with Graph Neural Networks} (GNN)~\cite{hamilton2020graph}. GNNs generalize convolutions to non-Euclidean data, and thus allow us to model large-scale global connections as edges of a graph. We visualize this key modeling difference in~\cref{fig:teaser}. 
Furthermore, GNNs can enhance model interpretability, given that domain knowledge can be encoded into the graph connection structure, or, if using an adaptive graph structure, we can analyze the learned edges.
Our proposed model also requires 12 times less models than a state-of-the-art deep learning-based approach~\cite{CNN_ENSO}, which requires a separate model for each target season (and each number of lead months).
Lastly, we note that GNNs are more efficient than recurrent neural networks and LSTMs~\cite{STGCN}, which are often used in ENSO forecasting models~\cite{DLENSO, climateNetworks_and_LSTM}, as well as significantly more efficient than dynamical models, which are compute- and resource-intensive.

Motivated by these modeling advantages that GNNs enjoy over other deep learning architectures, we can summarize our key contributions as follows:
\begin{itemize}
    \item 
    We propose the first application of GNNs to long range and seasonal forecasting.
    \item
    Building upon established previous research we develop and open-source \graphino\footnote{Code is available at \texttt{\url{https://github.com/salvaRC/Graphino}}}: a flexible graph convolutional network architecture for long range forecasting applications in the climate and earth sciences.
    \item
    We introduce a novel graph structure learning module, which makes our model applicable even without a pre-defined connectivity structure.
    \item
    We show that our model is competitive to state-of-the-art statistical and dynamical ENSO forecasting systems, and \emph{outperforms them for forecasts of up to six months}.
    \item
    We exploit our model's interpretability, to show how it learns sensible connections that are consistent with existing theories on ENSO dynamics predictability.
\end{itemize}


\begin{figure*}
  \centering
  \begin{subfigure}[b]{0.48\textwidth}  
    \includegraphics[width=1\linewidth,height=0.17\textheight]{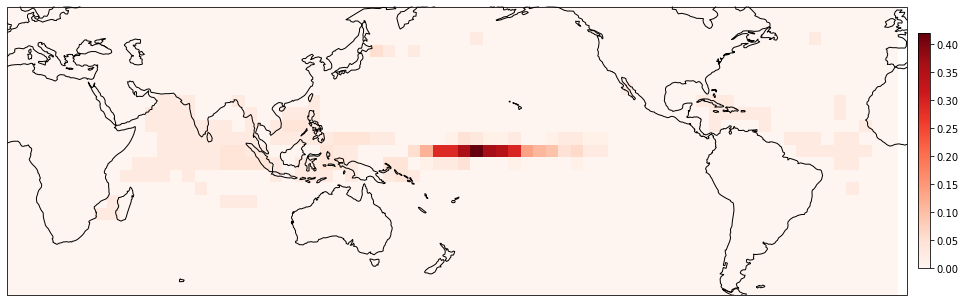}
      \caption{1 lead month}
      \label{fig:edge_heat_1lead}
  \end{subfigure}
  \begin{subfigure}[b]{0.48\textwidth}  
      \includegraphics[width=1\linewidth,height=0.17\textheight]{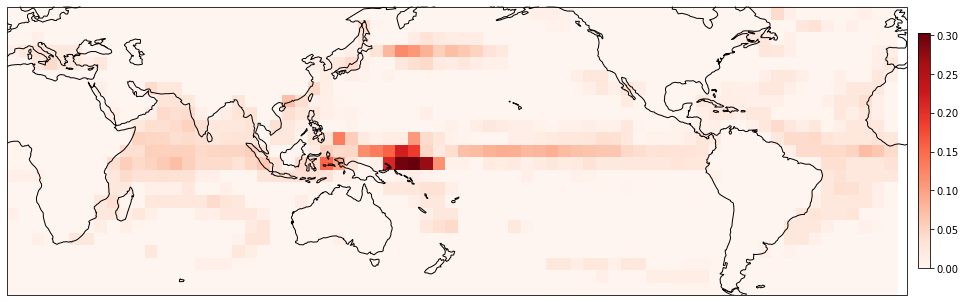}
      \caption{3 lead months}
      \label{fig:edge_heat_3lead}
  \end{subfigure}
  \begin{subfigure}[b]{0.48\textwidth}  
      \includegraphics[width=1\linewidth,height=0.17\textheight]{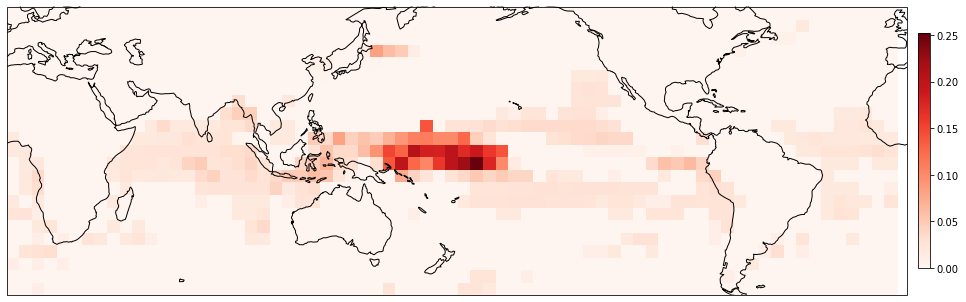}
      \caption{6 lead months}
      \label{fig:edge_heat_6lead}
  \end{subfigure}
  \begin{subfigure}[b]{0.48\textwidth}  
      \includegraphics[width=1\linewidth,height=0.17\textheight]{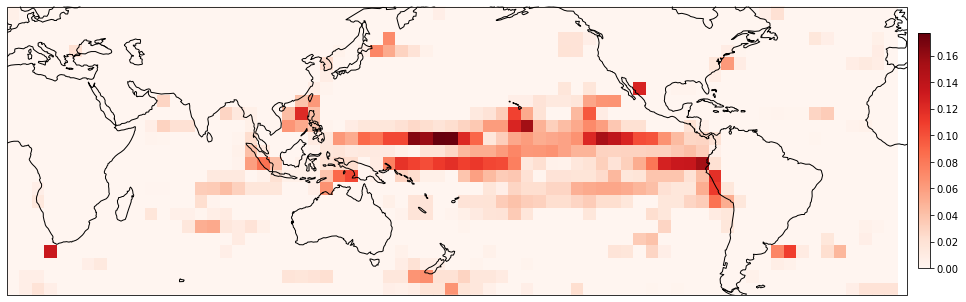}
      \caption{9 lead months}
      \label{fig:edge_heat_9lead}
  \end{subfigure}
  \begin{subfigure}[b]{0.48\textwidth}  
      \includegraphics[width=1\linewidth,height=0.17\textheight]{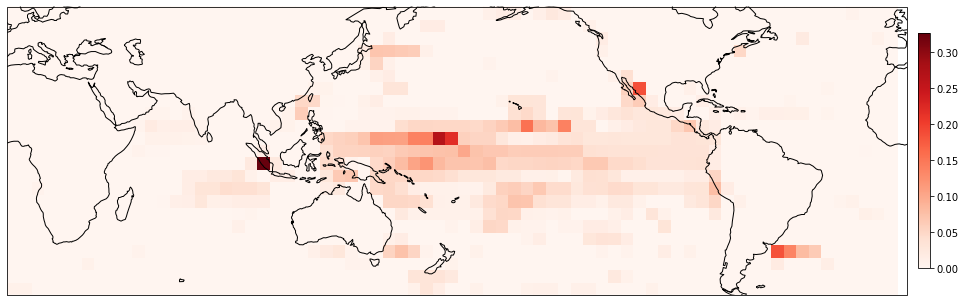}
      \caption{12 lead months}
      \label{fig:edge_heat_12lead}
  \end{subfigure}
  \begin{subfigure}[b]{0.48\textwidth}  
      \includegraphics[width=1\linewidth,height=0.17\textheight]{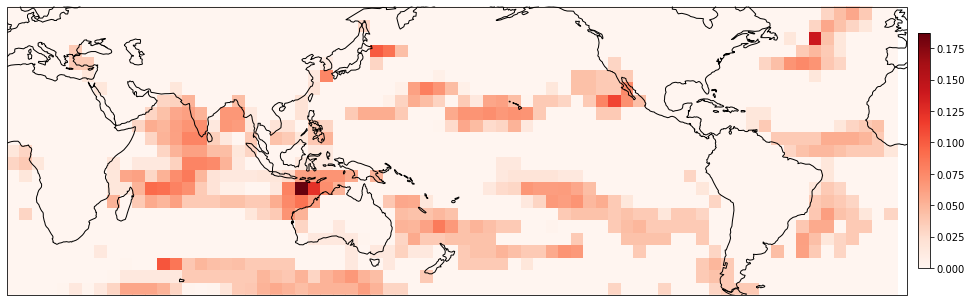}
      \caption{23 lead months}
      \label{fig:edge_heat_23lead}
  \end{subfigure}
  
  \caption{
  The learned world connectivity structure makes our proposed model \graphino$\!$ more interpretable than other black-box statistical models, while retaining a high predictive skill.
  To visualize the learned connectivity, we plot the eigenvector centrality of each node as a heatmap. It measures the influence of a node on the learned graph. Nodes with the highest importance can be seen in or near the ONI region for 1 lead month, while becoming more global with more lead months, as expected. Interestingly, our 23-lead model in \cref{fig:edge_heat_23lead} achieves a high correlation skill of $0.408$, which is considerably better than our ensemble and the main ensemble model from \cite{CNN_ENSO}.  Please refer to \cref{sec:Interpretation} for a more detailed discussion of the learned connectivity and its benefits.
  }
  \label{fig:edge_heat}
\end{figure*}
\renewcommand{\thesection}{\Roman{section}} 
\section{Background}
Methods to forecast ENSO can be broadly classified into dynamical and statistical systems \cite{ENSO_review, ENSO_predictability, ENS=_predictability2)}. The former are based on physical processes/climate models (e.g. atmosphere–ocean coupled models)~\cite{AModelENSO, palmer2004development, saha2010ncep}, while the latter are data-driven, including machine learning (ML)-based approaches. 

The presence of an ENSO event is commonly measured via the running mean over $k$ months of sea surface temperature anomalies (SSTA) over the Oceanic Niño Index (ONI, $k=3$) region (5N-5S, 120-170W), also known as the Niño3.4 index region ($k=5$).

\renewcommand{\thesection}{\Roman{section}} 
\section{Related Work} 
\subsection{Machine Learning for ENSO forecasting} $\;$ Recently, deep learning was successfully used to forecast ENSO 1\unit{yr} ahead \cite{TemporalCNN_ENSO} as well as with a lead time of up to $1.5\unit{yrs}$~\cite{CNN_ENSO}, thus out-performing state-of-the-art dynamical methods. 
Both project the Oceanic Niño Index (ONI) for various lead times. The former only use the ONI index time series as input of a temporal Convolutional Neural Network (CNN), while the latter feed sea surface temperature (SST) and heat content anomaly maps data to a CNN. 
We note that the predictive skill of the model in \cite{TemporalCNN_ENSO} can be mostly attributed to the use of a denoising method (EEMD \cite{EEMD}), which is contentious since the smoothing process may be transferring information from the future (i.e. test set) to the past \cite{EEMD_Controversial}.

Most statistical methods can only predict the single-valued index, an averaged metric over SST anomalies that does not convey zonal information. A notable exception, makes use of an encoder-decoder 
approach \cite{DLENSO}. An overview over other machine learning methods used to project ENSO is given in \cite{ML_for_ENSO}.
\subsection{Climate networks} $\;$ In climate networks \cite{ClimateNetworks}, which stem from the field of complex networks, each grid cell of a climate dataset is considered a network node and edges between pairs of nodes are set up using a similarity measure.
They have been used to detect and characterize SST teleconnections
~\cite{agarwal2019network} and to successfully forecast the presence of ENSO $1\unit{yr}$ prior~\cite{ENSO_EarlyForecast_with_climateNetworks}. The latter exploits the observation that, a year before an ENSO event, a large-scale cooperative mode seems to link the equatorial Pacific corridor (“El Niño basin”) and the rest of the Pacific ocean~\cite{ENSO_EarlyForecast_with_climateNetworks}.
Our GNN approach for ENSO forecasting builds on the climate network’s precedent of describing climate as a network of nodes related by
non-local connections. 

\subsection{Graph neural networks} $\;$
In the past years, GNNs have surged as a popular sub-area of research within machine learning~\cite{GNN_review}. Interestingly, they have scarcely been used in earth and atmospheric sciences. A few applications use them for earthquake source detection \cite{van2020Seisms_and_GNNs}, power outage prediction \cite{owerko2018Outages_and_GNNs} and wind-farm power estimation \cite{park2019Wind_power_and_GNNs}. The representation of data as a graph, however, makes GNNs a very promising candidate to learn distant relationships in ENSO forecasting. This work is the first to explore the performance of GNNs for seasonal forecasting.
As in the CNN-based work~\cite{CNN_ENSO}, we do not explicitly model temporal relationships in the present paper, but instead build upon the standard graph convolutional network architecture~\cite{kipf2017semi}.
A natural extension would therefore be to explicitly model the temporal patterns, e.g. with spatiotemporal GNNs that have already been extensively applied to traffic forecasting~\cite{STGCN, MTGNN, Syncronous_STGCN, WaveNet, adaptive_rec_conv_gnn}.

\begin{table*}
\caption{Our proposed GNN model, \graphino, outperforms the deep learning-based model, CNN~\cite{CNN_ENSO}, and the dynamical model, SINTEX-F~\cite{SintexF}, in all-season correlation skill for forecasts up to six months. 
Our model also outperforms other dynamical models from the North-American Multi-Model Ensemble project which were shown to perform at best as good as the proposed CNN in \cite{CNN_ENSO}.
The correlation skill was measured on the held-out GODAS test set (1984-2017) with the same setup as~\cite{CNN_ENSO}. 
The skill for SINTEX-F was generously estimated from Fig. 2a in~\cite{CNN_ENSO}. 
}
    \label{table:results_cnn}
    \centering
    \begin{tabular}{@{} *{16}l @{}}
    \toprule
     Model     & $n=1$ & $n=2$ & $n=3$ & $n=4$ & $n=5$ &
                 $n=6$ & $n=9$ & $n=12$ & $n=23$\\ 
    \midrule
    \text{SINTEX-F}~\cite{SintexF} & 
        0.895 & 0.89 & 0.84 & 0.805 & 0.78 & 
        0.74 & 0.62 & 0.51 & 0.315 \\
    \text{CNN}~\cite{CNN_ENSO} & 
        0.9423 & 0.9158 & 0.8761 & 0.8320 & 0.7983 & 
        0.7616 & \first{0.7133} & \first{0.6515} & 0.2870 \\
    \graphino & 
        \first{0.9747} & \first{0.9461} & \first{0.9170} & \first{0.8742} & \first{0.8226} &
        \first{0.7800} & 0.6313 & 0.5755 & \first{0.3363} \\
    \bottomrule
    \end{tabular}
\end{table*}
\section{Methods}
\label{sec:model_architecture}
To see how we can map general climate datasets into a problem appropriate for a GNN, we note that these datasets are usually gridded. Hence, the grid cells (i.e. geographical locations) can be naturally mapped to the nodes of a GNN. The graph's edges, which model the flow of information between nodes, are the main argument in favor of a GNN approach. Edges can be chosen based on domain expertise or on edges analyzed in climate networks research, or they can be jointly learned with the target forecasting task. The explicit modeling of interdependencies based on domain expertise, or the GNN's choice of meaningful edges (e.g. well known patterns or teleconnections), greatly enhances the model's interpretability. 
In this work, we propose a novel graph structure learning module to jointly learn the connectivity structure, and forecast the ONI.
In the following, we present the formal setup of our approach.

\subsection{Problem formulation}
Let $\mathcal{V} = \{V_1, V_2,\dots, V_N\}$ be the nodes of the graph $G=(\mathcal{V}, \mathcal{E})$, where each node $V_i$ is a grid cell of a (flattened) climate dataset, defined by its latitude and longitude.
We will later define the set of edges, $\mathcal{E}$. 
For each time step, $t \in \{1,\dots,T\}$, we associate with each such location a feature vector $\mathbf{V}_i^{(t)} \in \R^D$ of $D$ climate variables.
For time step $t$ let $\mathbf{X}_t = \{\mathbf{V}_1^{(t)}, \dots, \mathbf{V}_N^{(t)} \} \in \R^{N \times D}$ be a snapshot measurement over all locations.
Given a temporal sequence of such climate measurements $ \mathbf{X} = \{\mathbf{X}_{t_1}, \dots, \mathbf{X}_{t_w} \} \in \mathcal{X}$, our goal is to forecast the ONI index $Y=Y_{t_{w+h}}\in \R$
with $h$ months of anticipation and the window size, $w$, i.e. the number of timesteps used for prediction.
In this paper, we do not explicitly model the temporal component, and instead simply concatenate the $w$ temporal sequence elements $\mathbf{X}_{t_i} \in \R^{N \times D}$ to a single representation matrix $\mathbf{X} \in \R^{N \times wD}$.

Our dataset then consists of a timeseries of such pairs $(\mathbf{X}, Y)$ and our goal is to learn a neural network model $f_\phi: \mathcal{X} \rightarrow \R$
, parameterized by $\phi$. We do so by minimizing an appropriate loss function, here the mean squared error between the predicted and true ONI index.
We note that simple, but promising, extensions to this basic setup include multi-step forecasting, and forecasting of multiple values (e.g. multiple zonal SSTAs).

To finalize our graph's definition, we also need to define the set of edges, $\mathcal{E}$, that encodes the connectivity structure between geographical locations. This can be done through an adjacency matrix $\mathbf{A}\in \{0, 1\}^{N\times N}$, where $\mathbf{A}_{i,j}$ equals $1$ when there exists an edge from node $V_i$ to node $V_j$, and 0 otherwise. To mirror grid-convolutions from CNNs we could choose them according to geographical neighborhood, i.e. by connecting adjacent grid cells of the climate dataset.
This would, however, seriously limit the predictive power of our model, as we show in \cref{sec:ablations}, since ENSO is inherently a large-scale phenomenon.
In this paper, we choose to view the edge structure as learnable,  jointly with the model's parameters $\phi$. This makes it possible to inspect the learned adjacency matrix during, and after 
model training, to validate whether it is sensible.
We now introduce our structure learning component in detail.

\subsection{Graph structure learning module}
We propose a directed, end-to-end learnable edge structure represented by a continuous adjacency matrix, $\mathbf{A}\in [0, 1]^{N\times N}$, where each entry, $\mathbf{A}_{i,j}$, can be now interpreted as a weighted connection from node $V_i$ to node $V_j$.
In the following we use tildes for variables only occurring in our structure learning module (e.g. $\tilde{\mathbf{X}}$), that should not be confused with similar, but unrelated counterparts without tilde used by the GCN module (e.g. $\mathbf{X}$).
Concretely, we make use of static node representations, $\tilde{\mathbf{X}} \in \R^{N\times \tilde d_1}$, to compute
\begin{align}
    \mathbf{M}_1 
     &= 
     \text{tanh}\brackets{\alpha_1 \tilde{\mathbf{X}}\tilde{\mathbf{W}}_1} \in \R^{N\times \tilde d_2}, 
     \label{eq:m1} \\
    \mathbf{M}_2
    &= 
    \text{tanh}\brackets{\alpha_1 \tilde{\mathbf{X}}\tilde{\mathbf{W}}_2}
    \in \R^{N\times \tilde d_2},
    \label{eq:m2} \\
    \mathbf{A}
    &= 
    \text{sigmoid}\brackets{\alpha_2 \mathbf{M}_1\mathbf{M}_2^T} \in [0, 1]^{N\times N},
    \label{eq:adp} 
\end{align}
where $\tilde{\mathbf{W}}_1, \tilde{\mathbf{W}}_2 \in \R^{\tilde d_1\times \tilde d_2}$ are learnable parameters, and $\alpha_1$ and $\alpha_2$ are hyperparameters. The smaller $\alpha_1$ is, the more distinct values are generated, while a high $\alpha_2 > 1$ leads to more confident scores on whether there is an edge or not (i.e. it discourages values close to $0.5$).
In a final step, we set all but the largest $e$ edge values $\mathbf{A}_{i, j}$
to $0$ in order to enforce a sparse connectivity structure.
This module is inspired by the unidirectional module from \cite{MTGNN}.
The key differences are 1) We allow for bi-directional edges, while uni-directional edges are enforced in \cite{MTGNN}; 2) We set an upper limit, $e$, on the \emph{total} number of edges, instead of fixing a maximum number of neighbors for each node.
Note that while these differences are subtle, they are absolutely key for a better performance, and it lends itself for a better interpretation of the learned connectivity structure. We discuss these advantages in more detail in~\cref{sec:ablations}. 
As is standard practice, we add self-loops to the graph, so as to preserve node information, by letting the diagonal of $\mathbf{A}$ be non-zero.

\subsection{Graph Neural Network}
The problem of forecasting the ONI can be framed as a graph regression problem. As such, any GNN can be used to model this task. For this work, we build upon the popular graph convolutional neural network (GCN) architecture \cite{kipf2017semi}. At each GCN layer $l$, node embeddings $\mathbf{Z}^{(l)}_i$ are generated for each node $V_i$. The node embeddings $\mathbf{Z}^{(l)}_i$ for node $V_i$ are aggregated from the previous-layer embeddings of its neighbors: $\{\mathbf{Z}^{(l-1)}_j: A_{ij} \neq 0\}$. Thus, in the deeper layers, information from more distant nodes propagates to each node embedding. This process is therefore also called message-passing.
Mathematically the node embeddings $\mathbf{Z}^{(l)}$ of the $l$-th graph convolutional layer can be written as: 
\begin{equation}
    \mathbf{Z}^{(l)} = \sigma \brackets{\mathbf{A} \mathbf{Z}^{(l-1)}\mathbf{W}^{(l)}} \in \R^{N\times D_l}, \label{eq:gcn_layer}
\end{equation}
where $\mathbf{Z}^{(l-1)} \in \R^{N\times D_{l-1}}$ are the node embeddings of the previous layer (with the first layer, $\mathbf{Z}^{(0)} = \mathbf{X})$, and the activation function, $\sigma$.

In the standard case with a discrete $A_{i,j} \in \{0, 1\}$ the aggregation $\mathbf{A} \mathbf{Z}^{(l-1)}$ is a simple sum, while with our continuous graph structure learning formulation, $A_{i,j} \in [0, 1]$, it becomes a weighted sum.

We then use the output of the last layer $L$ to pool a graph embedding $\mathbf{g} \in \R^{D_L}$ by aggregating the node embeddings:
\begin{equation}
    \mathbf{g} = \text{Aggregate}\brackets{\mathbf{Z}^{(L)}_1, \dots, \mathbf{Z}^{(L)}_N},
\end{equation}
where the aggregation function can be, e.g., a mean or sum over the node embeddings.  In a final step, the graph representation $\mathbf{g}$ is used as input to a fully connected multi-layer perceptron, to forecast our estimate of the ONI index: $\hat Y = \text{MLP}\brackets{\mathbf{g}}$.

In practice, we use a more complex formulation with jumping knowledge and residual connections, which have been shown to increase performance in a wide variety of tasks~\cite{jumpingKnowledge}. Jumping knowledge connections, means that the node embeddings from the intermediate GCN layers are concatenated to the final one from layer $L$~\cite{jumpingKnowledge}. Please refer to the Appendix for the full mathematical formulation.
Further, while the original graph convolution normalizes the input node embeddings by the in-degree, we have found that replacing it with batch normalization over the feature dimension gives better results.

\begin{figure*}
\begin{subfigure}{.32\textwidth}
  \centering
  \includegraphics[width=1\linewidth]{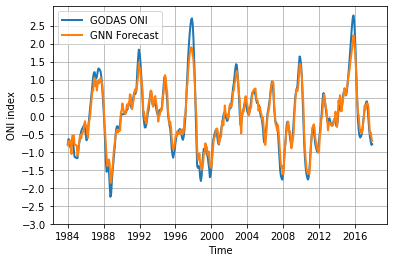}
  \caption{1 lead months, RMSE $=0.222$}
  \label{fig:1lead}
\end{subfigure}%
\begin{subfigure}{.32\textwidth}
  \centering
  \includegraphics[width=1\linewidth]{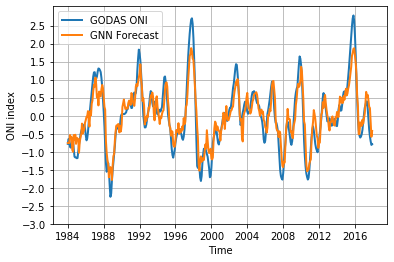}
  \caption{3 lead months, RMSE $=0.375$}
  \label{fig:3lead}
\end{subfigure}
\begin{subfigure}{.33\textwidth}
  \centering
  \includegraphics[width=1\linewidth]{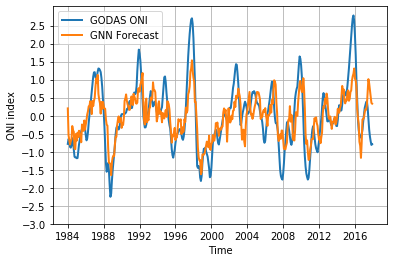}
  \caption{6 lead months, RMSE $=0.576$}
  \label{fig:6lead}
\end{subfigure}
  \caption{Our proposed graph neural network (orange) accurately predicts the Oceanic Ni\~no index (ONI) timeseries (blue). The high correlation skill of $0.97, 0.92$ and $0.78$ for the 1-, 3- and 6-month lead times confirms the accurate prediction of the trend.  While the predictions are strong for standard ENSO events, the GNN underpredicts the strength of the record-year el Ni\~no events in 1998 and 2016. 
  }
  \label{fig:timeseries}
\end{figure*}

\begin{figure*}[h]
 \centering
 \begin{subfigure}[b]{0.98\textwidth}  
      \includegraphics[width=1\linewidth]
      {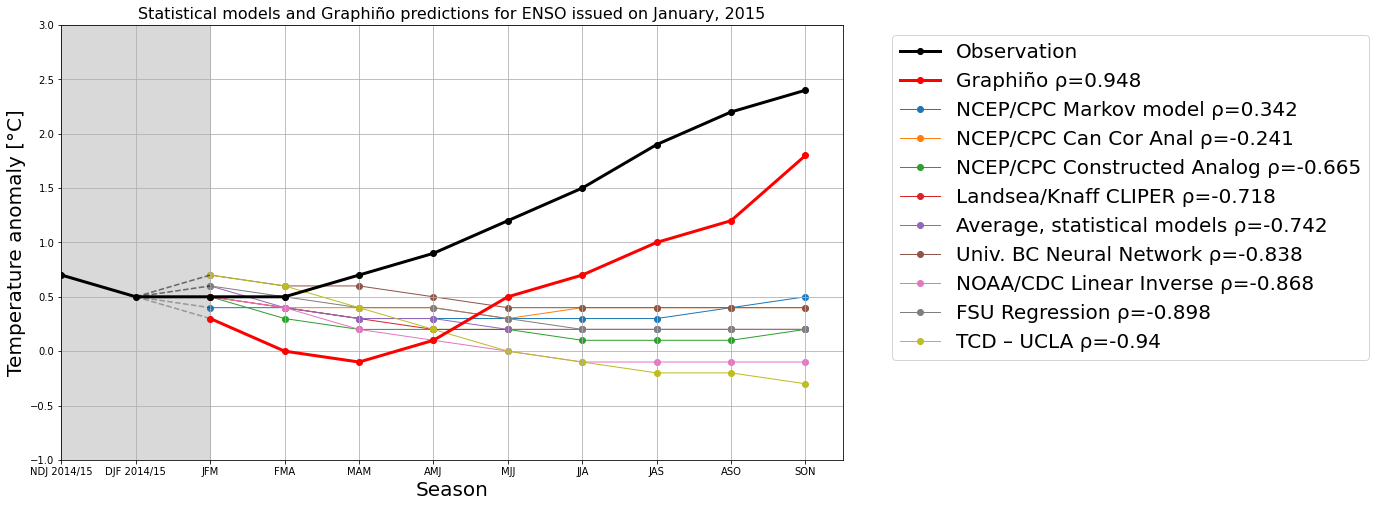}
      \label{fig:vsStat}
  \end{subfigure}
  \begin{subfigure}[b]{0.98\textwidth}  
      \includegraphics[width=1\linewidth]
      {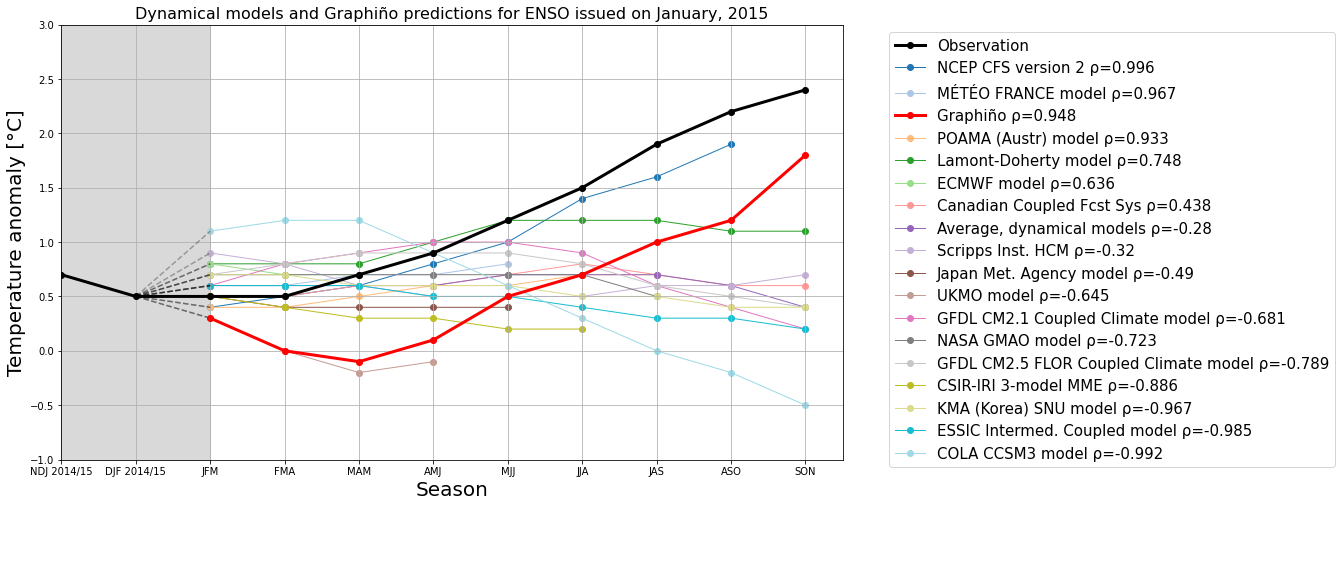}
     \label{fig:vsDyn}
   \end{subfigure}
   \vspace{-8mm}
   \caption[graphino plume]{\graphino \;(red) outperforms all statistical models (top) and most dynamical models (bottom) in the prediction of the extreme El Ni{\~n}o year, 2015.
    The plume plot shows the ONI forecasts issued on January 2015. 
    The entries in legends are sorted by the correlation coefficient $\rho$ with the first record being the best one (highest $\rho$). The lines correspond to the predicted ONI for various seasons, e.g., JFM being the running average of SST anomalies in the ONI region for January, February and March. Dynamical and statistical model predictions were provided by the International Research Institute for Climate and Society (IRI)~\cite{IRI_2015},
    and the observations by the National Oceanic and Atmospheric Administration (NOAA)~\cite{NOAA_NCEP_plume_2021}.
}
\label{fig:gnn_vs_models} 
\end{figure*}


\section{Experiments}\label{sec:results}
\subsection{Experimental setup}
To validate the predictive skill of our GNN \graphino, we benchmark it on the same setup and in a fair comparison against the CNN and dynamical models from  \cite{CNN_ENSO}.
That is, as training set we use the same SODA reanalysis dataset (1871-1973), and climate model simulations from CMIP5, that were used in~\cite{CNN_ENSO}. The augmentation of the dataset with potentially noisy simulations is important in order to have a sample size suitable for deep learning methods~\cite{CNN_ENSO} (in total $30k$ samples, while SODA only has $1200$ samples). 
The GODAS dataset for the period of 1984 to 2017 serves as the held-out test set.
The datasets are used in a resolution of $5$ degrees, and only the geographical locations that lie within $55S-60N$ and $0-360W$ are used. This results in $N=1345$ nodes (after filtering out all terrestrial ones).
The features for each node are the sea surface temperature (SST) and heat content anomalies over $w=3$ months (i.e., $D=2, D_0=2\times3=6$). The prediction target is the ONI index for $h$ months ahead.
This is the exact same setup from \cite{CNN_ENSO}.

Differently than the CNN in~\cite{CNN_ENSO} however, we
\begin{itemize}
    \item 
    only run a single model for a given number of target lead months, whereas they run a separate model for each target season (i.e. one for DJF, another one for JFM, etc.). This makes our approach require 12 times fewer models.
    \item
    do not use the transfer learning technique, but instead train in a single training process both on CMIP5 simulations and the SODA dataset, since we observed massive overfitting to the small SODA dataset otherwise.
\end{itemize}
Also, note that our GNN is a more natural representation for the task, since we can simply filter out all terrestrial locations, whereas the inflexible CNN grid structure requires them to be present (with all features equal to zero).
Just as in \cite{CNN_ENSO}, we report the performance of an ensemble of size four. For more details, please refer to \cref{sec:ImplementationDetails}.

\subsection{Results}
\subsubsection{Average prediction}
We find that our proposed GNN model outperforms the state-of-the-art CNN of \cite{CNN_ENSO} for up to 6 lead months, as well as the competitive dynamical model SINTEX-F~\cite{SintexF} for all lead times, see \cref{table:results_cnn}.
Our model is able to significantly advance the state-of-the-art of machine learning for ENSO models in this seasonal forecasting range of up to six months. We hypothesize that the more rapid decrease in performance of our model for more than six lead months compared to the CNN model, can be attributed to the fact that our model also needs to learn the connectivity structure. This potentially leads to a higher sample complexity and makes the model more prone to overfitting.
The ONI time series forecasted by our model, for $n=1,3,6$ lead months, are plotted in \cref{fig:timeseries}.

Interestingly, we found that one of our ensemble members achieves a correlation coefficient of $0.408$ for $23$ lead months, which is very high given that the skill of both, our ensemble and the one from \cite{CNN_ENSO}, only achieve $0.34$ and $0.29$ respectively.
This indicates, that further research could potentially achieve skillful multi-year forecasts. We therefore include the learned connectivity structure of this model in \cref{fig:edge_heat_23lead}, as we believe that it may be of interest to the community.

\subsubsection{Rare event prediction}
\cref{fig:gnn_vs_models} compares \graphino$\!$ to established selected statistical and dynamical models predicting ENSO events for year 2015, during which occurred one of the strongest El Niño in past decades \cite{iskandar_how_2018}. \graphino$\!$ considerably outperforms all of the statistical, and a vast majority of the dynamical models, being notably better than the average of both of the groups, achieving a correlation coefficient of 0.948 across the year.

\subsection{Analysis of the connectivity structure learned by the GNN}
\label{sec:Interpretation}
Recall that in our experiments, we do not pre-define any edges between the nodes. Thus, the GNN model needs to learn suitable edges in addition to the actual forecasting task.
Note that while we associate each node with its geographical location defined by its latitude and longitude, the GCN model has no notion of spatiality.
That being said, we now analyze the learned connections and demonstrate their reasonableness, which emphasizes our argument that we can encode powerful and interpretable inductive biases into GNNs for seasonal forecasting. 

\subsubsection{Eigenvector centrality} $\;$
Since the number of learned edges and  nodes is too large to explicitly plot every learned connection in an informative way, we instead choose to analyze the assigned importance of each node, as measured by their eigenvector centrality.
This quantity measures the influence of a node on the graph -- a node with a high centrality score means that it is connected to many other nodes with high scores, and therefore plays a central role in the graph. 
In mathematical terms, the centrality score of a node, $i\in\{1,\dots,N\}$, is the $i$-th element of the eigenvector $\mathbf{v} \in R_+^N$ that corresponds to the largest eigenvalue $\lambda_\text{max}$ of the adjacency matrix. That is, it satisfies $\mathbf{Av} = \lambda_\text{max} \mathbf{v}$~\cite{Networks}.

\subsubsection{Interpretation of learned edges in terms of ENSO models} $\;$
In \cref{fig:edge_heat} we plot heatmaps of the computed eigenvector centrality scores of each node of our best performing model for lead times $h\in \{1, 3, 6, 9, 12, 23\}$. These show which nodes (in darker color) play a central role in the graph and are connected to other central nodes. High eigenvector centrality scores translates to having a high influence in the GCN since the node's information will spread more during message passing. 

 In order to interpret the spatial features of the edges in terms of physical processes we first need to understand the basic physics behind ENSO. El Ni\~no or La Ni\~na events are characterized by the Bjerknes feedback, which is a positive feedback between SST and wind anomalies \cite{bjerknes69_atmos_pacif}. Wind tends to rush towards (away from) regions of warm (cold) SST which pushes even more warm water towards (away from) that region. This further intensifies the wind, which further intensifies the SST anomaly. During EL Ni\~no (La Ni\~na) events warm (cold) SST anomalies form over the eastern Pacific thereby intensifying the wind towards the eastern Pacific. If the Bjerknes feedback is not interrupted, the eastern Pacific would keep on getting warm (cold). The theories of ENSO attempt to explain the processes that interrupt the feedback and cause transition from one phase to another. The most popular theories imagine the tropical Pacific as a self-sustaining oscillator. The Rossby and Kelvin waves \cite{matsuno66_quasi_geost, gill80_some}, two standard modes of variability observed in the ocean, play a crucial role in these models. The Rossby (Kelvin) wave propagates westward (eastward). Moreover both can either deepen or shoal the thermocline, which can be considered to be the boundary between the warm surface water and cold deeper water, thereby increasing or lowering the heat content in the ocean. These are called downwelling and upwelling waves, respectively. 
 
 The delayed oscillator mechanism \cite{suarezschopf88_delay_action} suggests that during a neutral phase of ENSO an anomalous eastward wind burst could set off an eastward downwelling Kelvin wave and a westward upwelling Rossby wave. The eastward Kelvin wave, the speed of which is well-known, carries the warm water from the western Pacific warm pool towards the east thereby heralding the onset of El Ni\~no. Similarly, an anomalous westward windburst triggers upwelling Kelvin and downwelling Rossby waves leading to a La Ni\~na in the winter. The Kelvin waves are generally set off in the boreal spring and reach the eastern Pacific by the winter of the same year. Thus, most of our short term predictability, for example \cref{fig:edge_heat_3lead} and \cref{fig:edge_heat_6lead}, can be explained by the propagation of these waves. Once these waves are set off in the spring it becomes highly likely that an El Ni\~no or a La Ni\~na event would occur in the winter. Dynamical models also do a good job of predicting the El Ni\~no/La Ni\~na once these waves are triggered. However, they struggle to predict whether these waves would be triggered in any given year leading to the spring predictability barrier \cite{balmaseda95_decad_season} (Fig. 4).
 
 Yet another theory, the western Pacific oscillator \cite{weisbergwang97_wester_pacif}, predicts that the SST and thermocline anomalies in the western Pacific warm pool trigger anomalous winds in the western Pacific setting off Kelvin waves towards the east. The SST and sea level pressure anomalies preceding the spring in the Ni\~no5 (\(120^{\circ}–140^{\circ}\)E, \(5^{\circ}S–5^{\circ}\)N) and Ni\~no6 (\(140^{\circ}–160^{\circ}\)E, \(8^{\circ}–16^{\circ}\)N) region trigger eastward wind anomalies in the western Pacific. These wind anomalies are deemed important according to this theory. Significant 9-month lead predictability of our model seems to originate from this region (\cref{fig:edge_heat_9lead}) which shows that this mechanism provides a source of predictability for our model even before the spring. 
 
 The recharge-discharge oscillator \cite{jin97_equat_ocean} is yet another theory which suggests that during the warm phase anomalous eastward wind drives poleward transport of warm water thereby shoaling the thermocline and reducing the equatorial ocean heat content in the tropical Pacific. This leads to a transition to the cold phase, which is associated with anomalous westward winds. The westward winds then drive warm water towards the tropical Pacific performing what is known as the recharge phase. The signature of the discharge (recharge) process where warm water is transported away from (towards) the equator is evident in figures \cref{fig:edge_heat_9lead} and \cref{fig:edge_heat_12lead}.

For two year lead in figure \cref{fig:edge_heat_23lead} we see hotspots of connectivity in the tropical Indian and Atlantic oceans. Warm SST over Indian and Atlantic ocean strengthens the trade winds from the Paicific thereby leading to weak El Ni\~nos or prolonged La Ni\~na-like conditions \cite{dongmcphaden18_unusual_indian, li15_atlan, luo12_indian_ocean}. More recently, a unified oscillator theory has been proposed \cite{wang01_unified_oscil,wang18_enso} which merges the effects of all the above mentioned oscillators and suggests that all of them play a role in ENSO dynamics to varying extents. It is then encouraging to see that our GNN model is able to decipher patterns of variability that can be linked to multiple known theories of ENSO.

\subsection{Connectivity structure ablation}
\label{sec:ablations}
To validate the advantage of modeling distant interdependencies in our GNN with help of our proposed connectivity structure learning module, we run the same main experiments but with a fixed adjacency matrix based on geographical neighborhood. Concretely, each node is connected to all nodes within a radius of $5$ degrees, i.e. a center node has $8$ neighbors. We then run the same GNN as in the main experiments, but without the structure learning module.
As expected, we find that incorporating information from distant parts of the world is critical. Our flexible GNN that can learn an arbitrary connectivity structure considerably outperforms the same GNN provided with a fixed, local connectivity structure only, see \cref{table:localGNN}.

Furthermore, we find that our proposed structure learning module significantly outperforms the structure learning module proposed in \cite{MTGNN}.
This can be attributed to the more appropriate inductive biases imposed by our proposed method. Concretely, \cite{MTGNN} enforces uni-directional connections, whereas we give the module the freedom to learn arbitrary connections. Importantly, they enforce each node of the $N$ nodes to have $k$ neighbors/connections, whereas we only impose an upper limit on the total number of connections (e.g. $kN$).
This is a better inductive bias for seasonal forecasting since we expect that some nodes (e.g. around the ONI region) will be considerably more important than others. Therefore, such nodes should be more connected than other nodes (i.e. should play a more central role in the GNN message passing) whose information is less/not important for the downstream ONI forecasting task.
\begin{table}[h!]
\caption{Incorporating geographical distant information is key for a strong performance. We report the correlation skill for $n$ lead months of the same GNN with 1) our structure learning module, 2) the structure learning module from \cite{MTGNN}, and 3) a fixed, local connectivity structure with edges based on spatial proximity (local).}
    \label{table:localGNN}
    \centering
    \begin{tabular}{@{} *{5}l @{}}
    \toprule
     Edge structure     & $n=1$ & $n=3$ & $n=6$ & $n=9$\\ 
    \midrule
    Local & 
        0.9063 & 0.7752 & 0.5946 & 0.4586  
         \\
    \cite{MTGNN} & 
      0.9117 & 0.8503 & 0.6439 & 0.4190 
     \\
    \graphino & 
        \first{0.9747} & \first{0.9170} & \first{0.7800} & \first{0.6313} 
        \\
    \bottomrule
    \end{tabular}
\end{table}

\subsection{Implementation details} 
\label{sec:ImplementationDetails}
As in \cite{CNN_ENSO}, we report the predictive skill of an ensemble of four models. 
Two of them are 2-layer GCNs with layer sizes of $250\times100$ and $250\times250$
. The other two are 3-layer GCNs with mean and sum pooling concatenated as the output of the graph representation. The dimensions are $200\times200\times200$, and $250\times250\times250$ respectively.
To avoid overfitting with the larger, more complex 3-layer GCNs, we apply a L2-weight decay of $10^{-4}$, and $10^{-3}$ respectively, while the 2-layer GCNs are trained with a L2 weight decay of $10^{-6}$ only.
Note that graph networks often perform best with few layers, differently than, e.g., CNNs.
All of the GCNs are followed by a two-layer MLP.
For both networks we use the ELU activation function~\cite{ELU}.
The batch size is $64$, and we use SGD with a learning rate of $0.005$ and Nesterov momentum of $0.9$ as the optimizer.
We do not use neither a learning rate scheduler nor dropout.
We report the held-out test performance on GODAS of the last checkpoint after 50 epochs of training.
As indicated before, we found that batch normalization over the feature/embedding dimension gave better results than the standard in-degree normalization proposed in \cite{kipf2017semi} for the GCN. The MLP uses batch normalization too.
We set the static node representations, $\tilde{\mathbf{X}}$ to be equal to the SODA timeseries of SST and heat content anomalies, plus the latitude and longitudes of the nodes. Note that when no static representations are available, $\tilde{\mathbf{X}}$ can be learnable embeddings as in~\cite{MTGNN}.
To mimic an average number of neighbors equal to the number of adjacent cells used in a 3x3 CNN filter, we choose a maximum number of edges $e=8N$.
We set $\alpha_1 = 0.1$ and $\alpha_2 = 2$.
Since \graphino$\!$ is flexible enough to support additional nodes, we add an ONI node represented by the averaged out SST and heat content anomalies over the ONI region for each time step.
In each optimization step (i.e. for each batch), we retrieve
the adjacency matrix $\mathbf{A}$ from our structure learning module. It is then used in the following GNN forward pass. We then jointly optimize for the parameters $\phi$ of our main model $f_\phi$, and the structure learning parameters $\tilde{\mathbf{W}}_1, \tilde{\mathbf{W}}_2$, based on the mean squared error between the predicted $\hat Y = f_\phi(\mathbf{A}, \mathbf{X})$ and true ONI $Y$.
Our 2-layer models run at around $40s/$epoch, while the 3-layer models require around $60s/$epoch on a 4xGeForce GTX 1080Ti GPU. The total training time is $33$ and $50$ min respectively. After the training, inferences are very fast, as is usual with neural networks.
\section{Discussion}

Our proposed GNN approach for forecasting ENSO outperforms dynamical models like the SINTEX-F~\cite{SintexF} (in \cite{CNN_ENSO}, the proposed CNN is also shown to outperform the North American Multi-Model Ensemble members), and is better or comparable to state-of-the-art machine learning models~\cite{CNN_ENSO}.
Furthermore, our methodology outperforms the aforementioned study for seasonal forecasts, indicating the potential for improved model performance for longer leads multi-year forecasts with the inclusion of more variables, 
or an extensive hyperparameter search.

Our proposed approach is easily applicable to other important complex weather and climate forecasting problems. Besides an increased predictive skill with our model, we expect that the connectivity structure encoded into our model, that is nicely interpretable in earth sciences applications, will be just as, if not more, valuable to the community.
Lastly, we believe that a very promising direction is to further improve the graph structure learning module, and use the learned connectivity structure to advance our current understanding of the predictability of ENSO.
Application of this ENSO model could have a significant impact on weather prediction and human preparation if leveraged as a tool by climate researchers to provide better as well as longer lead-time predictions.
This would also allow global populations to better prepare for the predicted climate and its effects on industry, agriculture, safety, and human quality of life.

A limitation of our model is the underestimation of the extreme ENSO events, as can be seen in Figure \ref{fig:timeseries}.
By definition, these events are rare, which makes it a hard task for an ML model to correctly predict them from the limited sample size. A promising research direction may therefore focus on skillfully forecasting these extreme ENSOs, e.g supported by a custom loss function.
\vspace{-1.8mm}
\section{Conclusion}
Our proposed \graphino $\!$ model, based on a GNN architecture and a graph connectivity learning module, outperforms state-of-the-art ENSO forecasting methods for up to 6 months lead time. Our work shows promising results for the use of a GNN architecture for ENSO forecasting and other atmospheric modeling purposes, while also enhancing the ML model interpretability. 
Future work using other relevant climate variables with GNNs and better architectures, e.g. that explicitly model the temporal axis, could further improve forecasting results as well as provide novel information regarding the relationship between global regions as represented by the learned connectivity structure. 
 
\hfill March 12, 2021



%

\newpage
\appendices
\section{Full mathematical formulation}
In this section we formally define the graph convolutional network (GCN) model in the full form that we use, with jumping knowledge and residual connections. Both of these methods aim at increasing the quality of the final node representations, and are motivated by the \emph{over-smoothing} issue to which GNNs are sensitive to. Over-smoothing refers to node representations becoming increasingly similar with the number of message passing iterations (i.e. layers in our GCN)~\cite{hamilton2020graph}.

Recall that the generated node embeddings $\mathbf{Z}^{(l)}$ of the $l$-th graph convolutional layer can be written as: 
\begin{equation}
    \mathbf{Z}^{(l)} = \sigma \brackets{\mathbf{A} \mathbf{Z}^{(l-1)}\mathbf{W}^{(l)}} \in \R^{N\times D_l}, 
\end{equation}
where $\mathbf{Z}^{(l-1)} \in \R^{N\times D_{l-1}}$ are the node embeddings of the previous layer, $\mathbf{Z}^{(0)} = \mathbf{X}$, and $\sigma$ is an activation function.
If we add a \emph{residual connection} to layer $l$, provided that $D_l = D_{l-1}$, this becomes:
\begin{equation}
    \mathbf{Z}^{(l)} = \sigma \brackets{\mathbf{A} \mathbf{Z}^{(l-1)}\mathbf{W}^{(l)}} + \mathbf{Z}^{(l-1)}.
\end{equation}
Recall that in the standard GCN setting, the final representation $\mathbf{Z}_i$ of each node $i$ is simply its node embedding $\mathbf{Z}^{(L)}_i\in\R^{D_L}$ at the last layer $L$.
\emph{Jumping knowledge} connections additionally incorporate the embeddings from intermediate layers~\cite{jumpingKnowledge}. That is, the final node embedding becomes the concatenation of the outputs of all layers:
\begin{equation}
    \mathbf{Z}_i = [{\mathbf{Z}_i^{(1)}}^T, \dots, {\mathbf{Z}^{(L)}_i}^T] \in \R^d,
\end{equation}
where $d=\sum_{l=1}^L D_l$.
We then use this final node representations to pool a graph embedding $\mathbf{g} \in \R^d$ by aggregating them:
\begin{equation}
    \mathbf{g} = \text{Aggregate}\brackets{\mathbf{Z}_1, \dots, \mathbf{Z}_N}.
\end{equation}
The aggregation function can be a simple average, that is:
\begin{equation}
        \mathbf{g} = \frac{1}{N}\sum_{i=1}^N \mathbf{Z}_i.
\end{equation}
We indeed use this approach for two of our ensemble members, while for the other two we additionally concatenate the sum over the node representations:
\begin{equation}
    \mathbf{g} = \left[ \sum_{i=1}^N \mathbf{Z}_i,  \frac{1}{N}\sum_{i=1}^N \mathbf{Z}_i \right] \in \R^{2d}.
\end{equation}
Other possible aggregation functions can be a $\max(\cdot)$, or an attention mechanism. The best such aggregation function often varies across the specific applications and datasets. Indeed, while we found that the simple mean gives consistently solid results, a more extensive study on the most appropriate graph pooling approaches for ENSO and seasonal forecasting is required. 

\clearpage

\section*{Acknowledgment}
We would like to thank the ProjectX organizing committee for motivating this work. We gratefully acknowledge the computational support by the Microsoft AI for Earth Grant. We would also like to thank Captain John Radovan for sharing his expertise regarding the current ENSO models and their global applications, Chris Hill for his guidance on machine learning for oceans, as well as Chen Wang for his guidance on GNN architecture.

The research was partially sponsored by the United States Air Force Research Laboratory and the United States Air Force Artificial Intelligence Accelerator and was accomplished under Cooperative Agreement Number FA8750-19-2-1000. The views and conclusions contained in this document are those of the authors and should not be interpreted as representing the official policies, either expressed or implied, of the United States Air Force or the U.S. Government. The U.S. Government is authorized to reproduce and distribute reprints for Government purposes notwithstanding any copyright notation herein.

\ifCLASSOPTIONcaptionsoff
  \newpage
\fi



%
\bibliographystyle{IEEEtran}
\bibliography{References}

%


\begin{IEEEbiography}[{\includegraphics[width=1in,height=1.25in,clip,keepaspectratio]{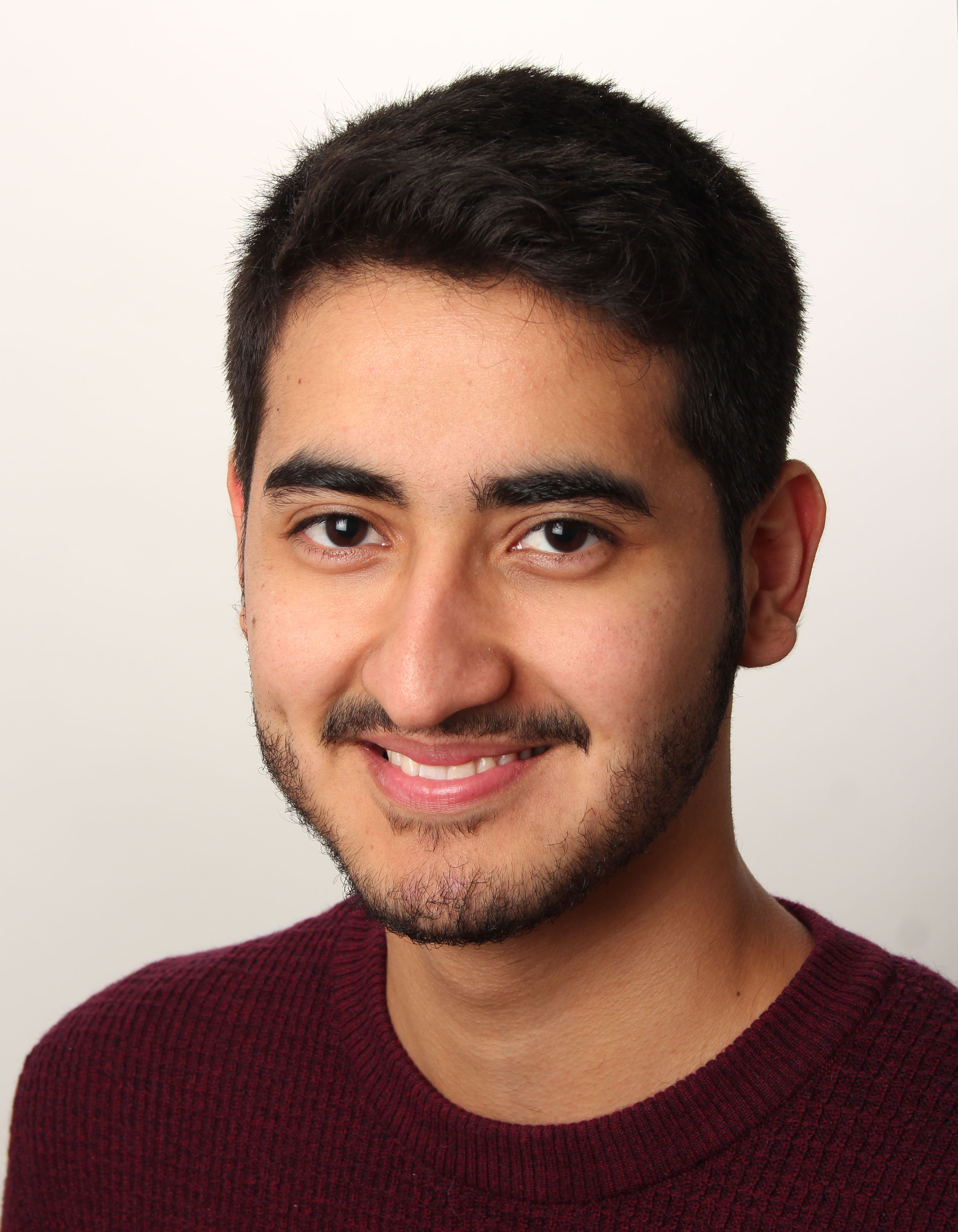}}]{Salva R{\"u}hling Cachay}
 is a computer science undergraduate student at the Technical University of Darmstadt, Germany. He has interned at Carnegie Mellon's University Auton Lab with Prof. Artur Dubrawski and Benedikt Boecking. He is currently pursuing a research internship with Prof. David Rolnick at Mila, Montreal.
 His research interests include weakly- and self-supervised learning as well impactful problems at the intersection of machine learning with the climate and earth sciences.
\end{IEEEbiography}

\begin{IEEEbiography}[{\includegraphics[width=1in,height=1.25in,clip,keepaspectratio]{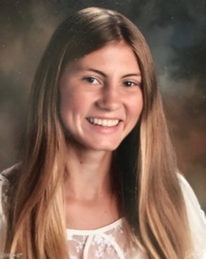}}]{Emma Erickson}
 is an undergraduate student studying electrical engineering at the University of Illinois at Urbana-Champaign. Previously she interned  under Prof. Artur Dubrawski and Robert Edman at Carnegie Mellon University, using machine learning to extract clinically relevant information from medical videos. Currently she researches active learning with the Computational Imaging Group at her home university under Prof. Minh Do and Corey Snyder. Her primary research interests lie at the intersection of signal processing, machine learning, and healthcare.
\end{IEEEbiography}

\begin{IEEEbiography}[{\includegraphics[width=1in,height=1.25in,clip,keepaspectratio]{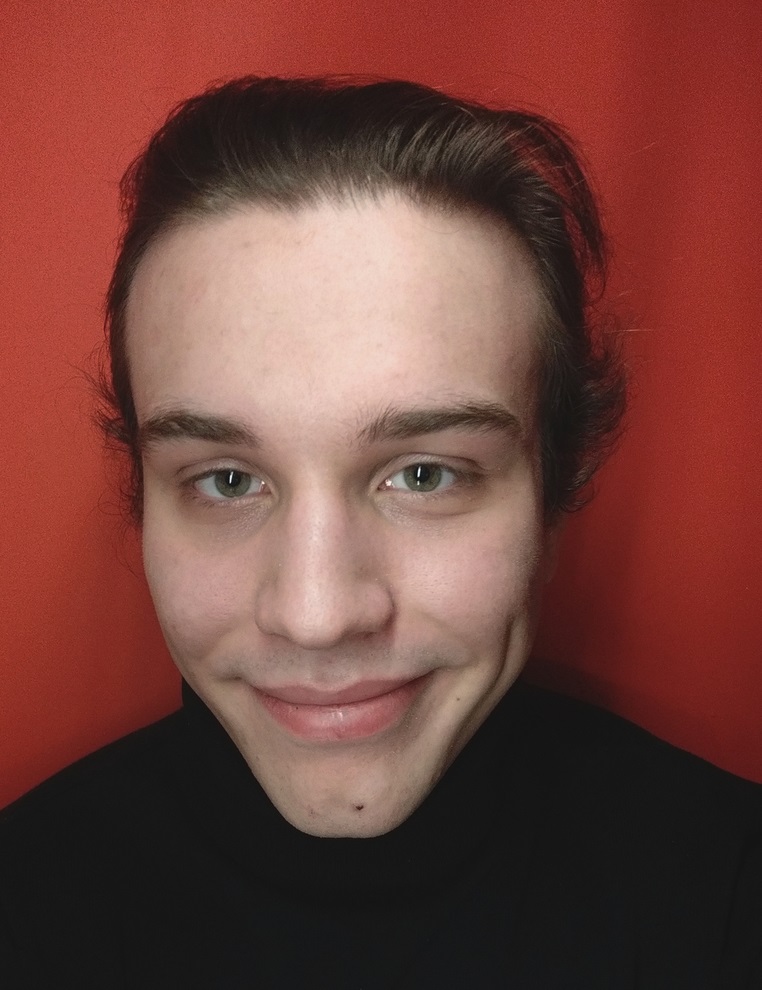}}]{Ernest Pokropek}
is a undergraduate student of computer science at Faculty of Electronics and Information Technology of Warsaw University of Technology, Poland. Co-founder of Polish student research group "FiberTeam", he is interested in applications of machine learning for various scientific domains, especially medicine and sensors, signal processing and feature extraction methodologies. 
\end{IEEEbiography}

\begin{IEEEbiography}[{\includegraphics[width=1in,height=1.25in,clip,keepaspectratio]{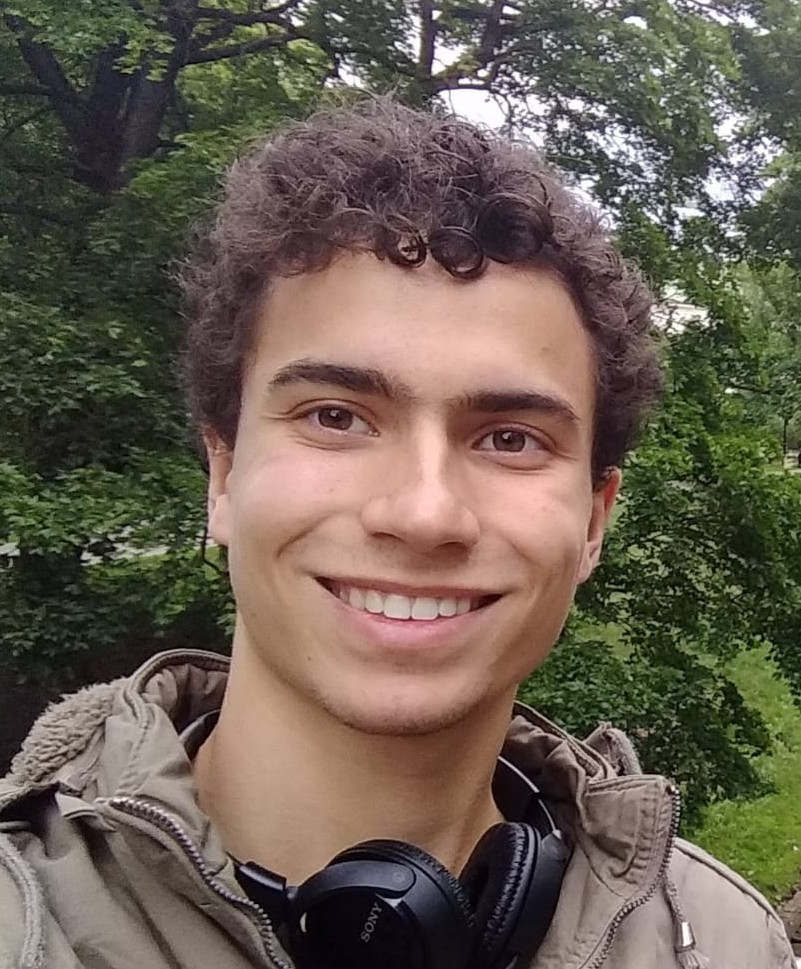}}]{Arthur Fender Coelho Bucker}
 is a mechatronics engineering undergraduate student at the University of S{\~a}o Paulo, Brazil, and is currently enrolled in a double degree program for master’s at the Technical University of Munich, Germany. His previous works at Carnegie Mellon’s  University AirLab address the fields of multi-robot motion planning and semantic control for autonomous aerial cinematography. Nowadays, his main research interests are in machine intelligence, AI applied to bio-robotics, and Brain-computer Interfaces.
\end{IEEEbiography}

\begin{IEEEbiography}[{\includegraphics[width=1in,height=1.25in,clip,keepaspectratio]{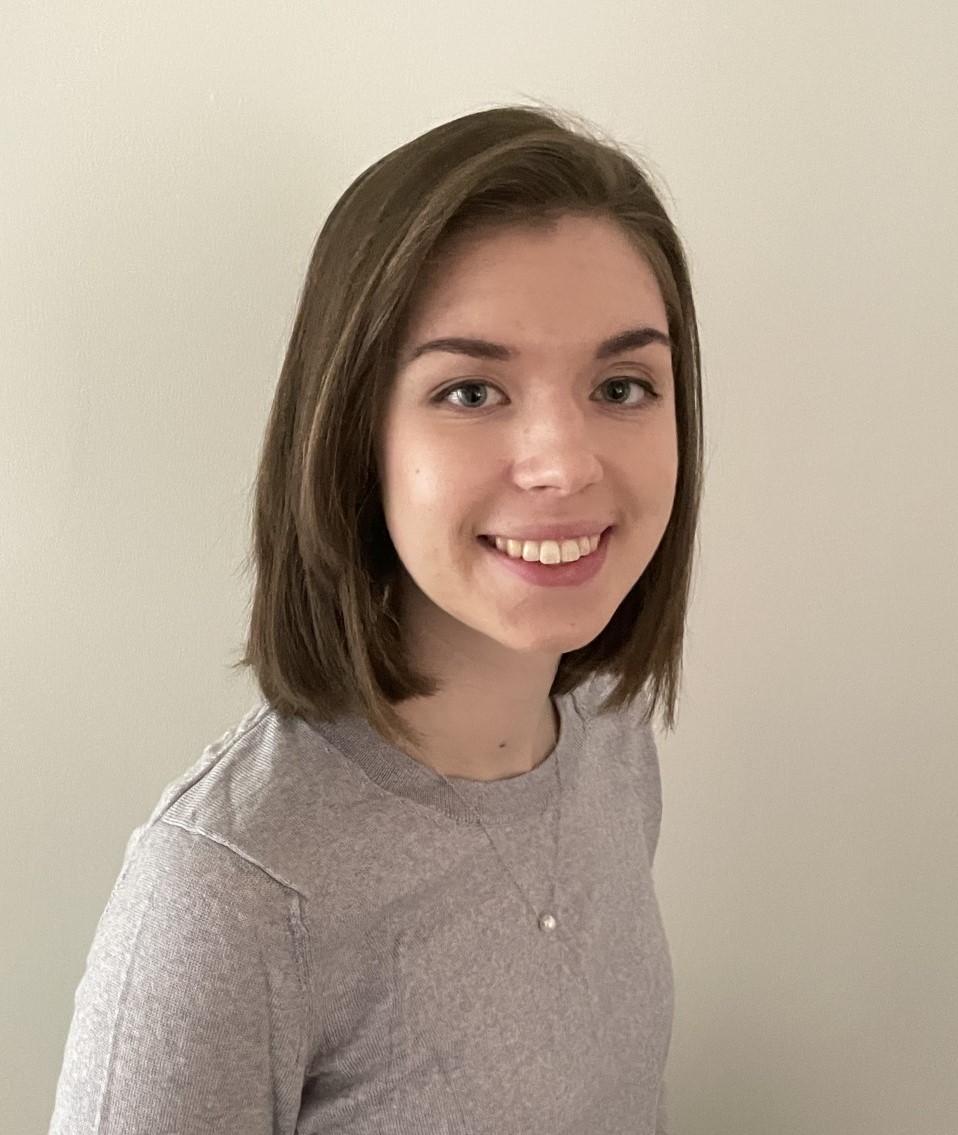}}]{Willa Potosnak} is a student in her 3rd year in the Biomedical Engineering Department at Duquesne University. She plans to continue her studies by pursuing a graduate degree in machine learning for medical applications and is interested in using advanced machine learning algorithms to improve predictive analytics. Currently, she is researching the use of machine learning to predict post-cardiac surgery renal failure using continuous intraoperative data.
\end{IEEEbiography}

\begin{IEEEbiographynophoto}{\textbf{Suyash Bire}} is a postdoctoral research associate at Earth, Atmospheric, and Planetary Sciences, MIT. He holds a Ph.D. in marine science from Stony Brook University with specializations in boundary current dynamics and interaction of large scale mean flow with turbulent eddies. He is currently working with Prof. John Marshall to explore the likely ocean circulation on icy moons in the solar system.
 His research interests include large-scale ocean circulation, eddy-mean flow interaction, boundary currents, and hydrothermal convection. He is also interested in exploring the applicability of machine learning techniques to climate and earth sciences.
\end{IEEEbiographynophoto}

\begin{IEEEbiography}[{\includegraphics[width=1in,height=1.25in,clip,keepaspectratio]{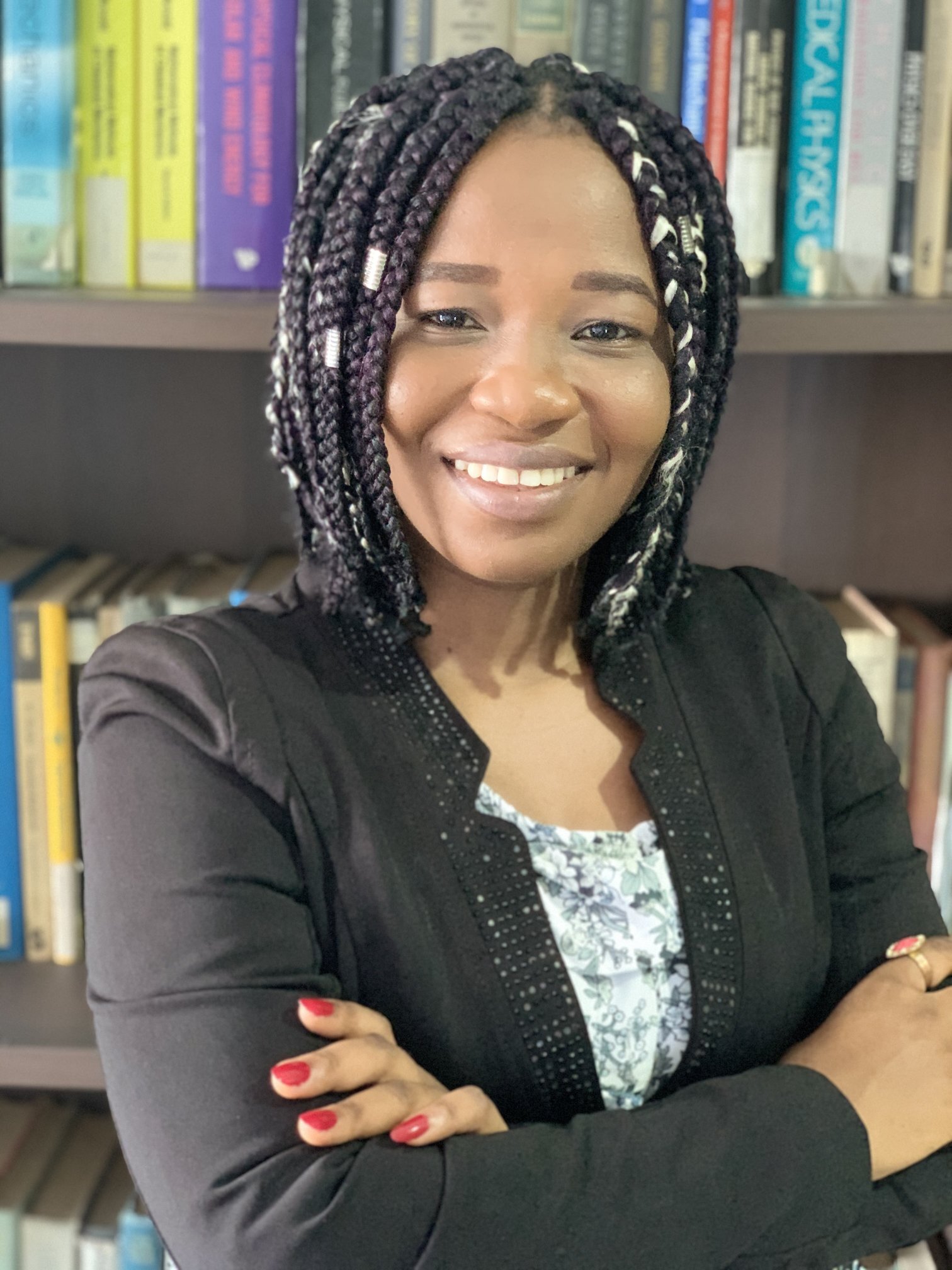}}]{Salomey Osei}
holds a Master of Philosophy in Applied Mathematics from the Kwame Nkrumah University of Science and Technology. She is the team lead of unsupervised methods for Ghana NLP and a co-organizer for the Women in Machine Learning and Data Science (WiMLDS) Accra chapter. Her research interest includes applications of machine learning, especially to finance and NLP. She is also passionate about mentoring students, especially females in STEM and her long term goal is to share her knowledge with others by lecturing.
\end{IEEEbiography}

\begin{IEEEbiography}[{\includegraphics[width=1in,height=1.25in,clip,keepaspectratio]{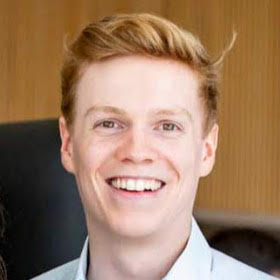}}]{Bj{\"o}rn L{\"u}tjens}
Bj{\"o}rn L{\"u}tjens is a PhD candidate at the Human Systems Laboratory, Department of Aeronautics and Astronautics, MIT. Together with Prof. Dava Newman, Dr. Cait Crawford, and Prof. Youssef Marzouk, he adapts  physics-informed neural networks to quantify the uncertainty in localized climate projections. This research positions him at the intersection of the physical/ecological sciences and machine learning with physics-informed neural networks, Bayesian deep learning, and robust optimization. He is also monitoring forest carbon from aerial imagery, which is supported by Microsoft, NASA, WWF, MIT PkG, MIT Legatum, and MIT Sandbox. He has previously obtained his Master's in Autonomous Systems, pioneering with Prof. Jon How safe and robust deep reinforcement learning techniques and holds a B.Sc. in Engineering Science from Technical University of Munich. 
\end{IEEEbiography}


\vfill


\end{document}